\documentclass[twoside,11pt]{article}
\usepackage[top=0.85in, bottom=0.85in, left=0.85in, right=0.85in]{geometry}

\usepackage{mathtools}
\usepackage{amsfonts}
\usepackage{amsmath}
\usepackage{amsthm}
\usepackage{amssymb}
\usepackage{booktabs}
\usepackage{xcolor}
\usepackage{balance}
\usepackage{algorithm,algorithmic}
\usepackage{multirow}
\usepackage{array}
\usepackage{tabularx}
\usepackage{amsbsy}

\usepackage[numbers]{natbib}
\def \esssup_#1{\underset{#1}{\mathrm{ess\,sup\, }}}
\def \essinf_#1{\underset{#1}{\mathrm{ess\,inf\, }}}
\def \argmax_#1{\underset{#1}{\mathrm{arg\,max\, }}}
\def \argmin_#1{\underset{#1}{\mathrm{arg\,min\, }}}

\def \beqs{\begin{eqnarray*}}
\def \enqs{\end{eqnarray*}}
\def \beq{\begin{eqnarray}}
\def \enq{\end{eqnarray}}

\def \N{\mathbb{N}}
\def \R{\mathbb{R}}

\def \Yc{{\mathcal Y}}

\def \Xc{{\mathcal X}}

\def \Sig{{\text{Sig}}}

\newtheorem{Theorem}{Theorem}

\newtheorem{Remark}{Remark}

\newtheorem{Definition}{Definition}
\newtheorem{Example}{Example}

\title{Transportation Marketplace Rate Forecast Using Signature Transform}

\author{
Haotian Gu
\thanks{Department of Industrial Engineering \& Operations Research,
       University of California, Berkeley, 
       Berkeley, CA 94720, USA \textbf{Email:} \{haotian\_gu, xinguo,  xinyu\_li\}@berkeley.edu}
\and
Xin Guo  
\footnotemark[1]
\thanks{Worldwide Operations Research Science, Amazon.com Inc., USA \textbf{Email:} \{xnguo, timojaco, philipka\}@amazon.com}
\and
Timothy L. Jacobs
  \footnotemark[2]
  \and
Philip Kaminsky
\footnotemark[2]
\and
Xinyu Li 
\footnotemark[1]
\footnote{A shorter version has appeared in KDD 2024.}
}

\begin{document}

\maketitle

\begin{abstract}
Freight transportation marketplace rates are typically challenging to forecast accurately.  In this work, we have developed a novel statistical technique based on signature transforms and have built a predictive and adaptive model to forecast these marketplace rates. Our technique is based on two key elements of the signature transform: one being its universal nonlinearity property, which linearizes the feature space and hence translates the forecasting problem into linear regression,  and the other being the signature kernel, which allows for comparing computationally efficiently similarities between time series data. Combined, it allows for efficient feature generation and precise identification of seasonality and regime switching in the forecasting process.

An algorithm based on our technique has been deployed by Amazon trucking operations, with far superior forecast accuracy and better interpretability versus commercially available industry models, even during the COVID-19 pandemic and the Ukraine conflict.  Furthermore, our technique is able to capture the influence of business cycles and the heterogeneity of the marketplace, improving prediction accuracy by more than fivefold, with an estimated annualized saving of \$50 million.

\end{abstract}

\section{Introduction}

\paragraph{Overview.} Linehaul freight transportation costs make up a significant portion of overall Amazon transportation costs.  To manage these costs, Amazon has developed a variety of tools to manage linehaul capacity mix and procurement.  One key input to all of these models is a  forecast of transportation freight marketplace rates, which however are notoriously difficult to forecast -- they are driven by a number of factors: the ever-changing network of tens of thousands of drivers, shippers of all sizes with a mix of occasional, seasonal, and regular demands, a huge set of brokers,  traditional and digital exchanges, and local, regional, national, and international economic factors of all kinds.  In addition, the transportation marketplace frequently goes through fundamental shifts -- either because of wars, pandemics, fuel prices, or due to shifting international trade patterns.  

Although Amazon has purchased externally created forecasts for some time, these forecasts are neither explainable nor sufficiently accurate to meet specific Amazon needs.  To address this challenge, we have built a forecasting model based on time series data to predict weekly freight marketplace rates for the North America market, at both the national and the regional levels. Our approach incorporates an innovative signature-based statistical technique capable of efficiently capturing significant fluctuations in transportation marketplace rates.

\paragraph{The key challenges in time series forecasting.} 
Time series data consists of sequential observations recorded over time and is ubiquitous: finance, economics, transportation, weather, and energy prices.    Given time series data, forecasting additional data points is critical for informed decision-making and process optimization in almost every organization and industry. 

Time series prediction models such as Autoregressive Integrated Moving Average (ARIMA) \citep{shumway2017arima} and Exponential Smoothing \citep{gardner2006exponential} assume that the time series are stationary, which is not the case for freight marketplace rates. Moreover, 
ARIMA has limited ability to capture seasonality and long-term trends \citep{kumar2015short}, and  Exponential Smoothing may be insufficient for abrupt changes or outliers and produce unstable forecasts \citep{taylor2004smooth}. Furthermore, these methods rely solely on historical data from the time series, which is inadequate in capturing the causal relation between the economic factors and the marketplace rates. Meanwhile,  machine learning algorithms such as Long Short-Term Memory Neural Networks \citep{yu2019review} and Gated Recurrent Units \citep{chung2014empirical}, though capable of capturing nonlinear relationship and complex patterns in time series data, will require substantially more training data which is not available in our case.

Indeed, one of the main challenges in analyzing time series data is their ever-changing statistical properties, due to factors including changes in business and economic cycles, shifts in policy, or changes in market conditions. In our case, the market itself has recently experienced shifts in \textit{regimes} and \textit{seasonality} \citep{hamilton1989new}, in terms of volatility, trends, and cyclical patterns, partly due to the COVID-19 pandemic and the Ukraine conflict.

\paragraph{Machine learning models and signature transform.}
Much of statistical learning theory relies on finding a feature map that embeds the data (for instance, samples of time series) into a high-dimensional feature space. Two requirements for an ideal feature map are \textit{universality}, meaning that non-linear functions of the data are approximated by linear functionals in the feature space; and \textit{characteristicness}, meaning that the expected value of the feature map characterizes the law of the random variable. It is shown that with the technique of the signature transform these two properties are in duality and therefore often equivalent \citep{simon2018kernel}. This is the primary inspiration for our proposed signature-based forecasting technique for our forecast models.

Originally introduced and studied in algebraic topology \citep{chen1954iterated, chen1957integration}, the \textit{signature transform}, sometimes referred to as the \textit{path signature} or simply \textit{signature}, has been further developed in rough path theory \citep{lyons2007differential, friz2010multidimensional},   introduced for financial applications \citep{lyons2014feature, arribas2018derivatives, lyons2019numerical, kalsi2020optimal} and machine learning \citep{yang2015chinese, xie2017learning, li2017lpsnet, bonnier2019deep, kidger2020signatory}, and most recently to time series data analysis \citep{morrill2020generalised,chevyrev2022signature,friz2020course}.  Given any continuous or discrete time series, their signature transform produces a vector of real-valued features that extract information such as order and area, and explicitly considers combinations of different channels. The signature of time series uniquely determines the time series, and does so in a computationally efficient way. Most importantly, every continuous function of a time series data may be asymptotically approximated by a linear functional of its signature. In other words, signature transform linearizes the otherwise complicated feature space, and thus is a powerful tool for feature generation and pattern identification in machine learning.

\paragraph{Our work.}

We 
propose a novel signature-based statistical technique for the time series forecasting problem. This is based on two key elements of the signature transform.
The first is the universal nonlinearity property of the signature transform,  which linearizes the feature space of the time series data and hence translates the forecasting problem into a linear regression. The second is the signature kernel which allows for computationally efficient comparison of similarities between time series data. Technically, this is to identify different ``signature feature maps'',  the statistical counterpart of identifying different distributions for a given time series data, albeit in the linearized feature space from the signature transform. 

Our approach starts by collecting data including a hundred of market supply and demand factors, and runs a correlation test between the marketplace rate and the factors to remove non-significant factors and to identify factors that may be colinear. We then exploit the universal nonlinearity property of signature transform to construct signature features as suitable candidates for the ``internal” features. To avoid issues of overfitting and co-linearity  between the signature
feature and the external factors, and to improve the forecast accuracy, we adopt the two-step LASSO for the regression analysis \citep{belloni2013least}.
Finally, this two-step LASSO is enhanced with adaptive weight using a signature kernel, which enables capturing changes in regimes or seasonality. 
Combined, this leads to our signature-based adaptive two-step LASSO approach. This novel signature-transform-based technique for data analysis allows for efficient feature generation and more precise identification of seasonality and regime switching embedded in the data. 

\paragraph{Implementation and real-time performance.} This signature-based adaptive two-step LASSO algorithm has been implemented for the trucking operations in Amazon since November 2022.  Performance analysis shows that our forecast model presents superior performance than commercially available forecast models, while providing significantly better interpretability. Despite the onset of COVID-19 and the Ukraine conflict,
it captures the influence of business cycles and heterogeneity of the marketplace, improves prediction accuracy by more than fivefold, and has an estimated annualized saving of approximately $\$50$ million.

 \section{Technical Background}

\subsection{Signatures of Continuous Paths}
We begin with the definition of signatures of continuous piecewise smooth paths.

\paragraph{Notation} Let $\mathbb{R}^{d_{1}} \otimes \mathbb{R}^{d_{2}} \otimes \cdots \otimes \mathbb{R}^{d_{n}}$ denote the space of all real tensors with shape $d_{1} \times d_{2} \times \cdots \times$ $d_{n}$. Define a binary operation called \textit{tensor product}, denoted by $\otimes$, which maps a tensor of shape $\left(d_{1}, \ldots, d_{n}\right)$ and a tensor of shape $\left(e_{1}, \ldots, e_{m}\right)$ to a tensor of shape $\left(d_{1}, \ldots, d_{n}, e_{1}, \ldots, e_{m}\right)$ via $\left(A_{i_{1}, \ldots, i_{n}}, B_{j_{1}, \ldots, j_{m}}\right) \mapsto A_{i_{1}, \ldots, i_{n}} B_{j_{1}, \ldots, j_{m}}$. When applied to two vectors, it reduces to the outer product. Let $\left(\mathbb{R}^{d}\right)^{\otimes k}=\mathbb{R}^{d} \otimes \cdots \otimes \mathbb{R}^{d}$, and $v^{\otimes k}=v \otimes \cdots \otimes v$ for $v \in \mathbb{R}^{d}$, in each case with $k-1$ many $\otimes$.

\begin{Definition}\label{def:signature_continuous}
Let $a < b \in \mathbb{R}$, and $X=\left(X^{1}, \ldots, X^{d}\right):[a, b] \rightarrow \mathbb{R}^{d}$ be a continuous piecewise smooth path. The signature of $X$ is then defined as the collection of iterated integrals
\begin{align*}
\operatorname{Sig}(X) &=\left(\int_{a<t_{1}<\cdots<t_{k}<b} \mathrm{~d} X_{t_{1}} \otimes \cdots \otimes \mathrm{d} X_{t_{k}}\right)_{k \geq 0} \notag\\
&=\left(\left(\int_{a<t_{1}<\cdots<t_{k}<b} \mathrm{~d} X_{t_{1}}^{i_{1}} \cdots \mathrm{d} X_{t_{k}}^{i_{k}}\right)_{1 \leq i_{1}, \ldots, i_{k} \leq d}\right)_{k \geq 0},
\end{align*}
where $\otimes$ denotes the tensor product, $\mathrm{d} X_{t}=\frac{\mathrm{d} X_{t}}{\mathrm{~d} t} \mathrm{~d} t$, and the $k=0$ term is taken to be $1 \in \mathbb{R}$. The truncated signature of depth $N$ of $X$ is defined as
\begin{equation}\label{eqn:sig_N}
    \operatorname{Sig}^{N}(X)=\left(\int_{a<t_{1}<\cdots<t_{k}<b} \mathrm{~d} X_{t_{1}} \otimes \cdots \otimes \mathrm{d} X_{t_{k}}\right)_{0 \leq k \leq N} .
\end{equation}
\end{Definition}

\begin{Remark}
The signature can be defined more generally on paths of bounded variation \citep{friz2020course}, but the above definition suffices for our purposes.
\end{Remark}

\begin{Example}\label{exp:sig_linear}
Suppose $X:[a, b] \rightarrow \mathbb{R}^{d}$ is the linear interpolation of two points $x, y \in \mathbb{R}^{d}$, so that $X_{t}=x+\frac{t-a}{b-a}(y-x)$. Then its signature is the collection of tensor products of its total increment:
\begin{equation*}
    \operatorname{Sig}(X)=\left(1, y-x, \frac{1}{2}(y-x)^{\otimes 2}, \frac{1}{6}(y-x)^{\otimes 3}, \ldots, \frac{1}{k !}(y-x)^{\otimes k}, \ldots\right),
\end{equation*}
which is independent of $a, b$.
\end{Example}

\begin{Example}\label{exp:sig_1d}
Suppose $X:[a, b] \rightarrow \mathbb{R}$ is a one-dimensional smooth path. Then its signature is the collection of powers of its total increment:
\begin{equation*}
\begin{aligned}
    \operatorname{Sig}(X)=\Big(&1, X(b)-X(a), \frac{1}{2}\left(X(b)-X(a)\right)^2, \frac{1}{6}\left(X(b)-X(a)\right)^{3}, \\
    &\quad \ldots, \frac{1}{k !}\left(X(b)-X(a)\right)^{k}, \ldots\Big),
\end{aligned}
\end{equation*}
which is independent of $X(t), t\in(a,b)$. Furthermore, when $X(t)$ is a random process, the expected signature
\begin{equation*}
\begin{aligned}
    \mathbb{E}\left[\operatorname{Sig}(X)\right]=\Big(&1, \mathbb{E}\left[X(b)-X(a)\right], \frac{1}{2}\mathbb{E}\left[\left(X(b)-X(a)\right)^2\right],  
    \\
    &\quad\ldots, \frac{1}{k !}\mathbb{E}\left[\left(X(b)-X(a)\right)^{k}\right], \ldots\Big),
\end{aligned}
\end{equation*}
whenever it exists, describes precisely the moments of $X(b)-X(a)$. Thus, for a high-dimensional stochastic process $X(t)$, the expected signature naturally forms the generalization of the moments of the process. In other words, for a stochastic process $X(t)$, its expected signature characterizes the law of $X(t)$ up to tree-like equivalence, as proved in \cite{chevyrev2016characteristic}.
\end{Example}

Example \ref{exp:sig_1d} shows that the signature for a one-dimensional path only depends on its total increment. In general, it implies that the signature of a path itself may not carry sufficient information to fully characterize the path. Nevertheless, this problem may be resolved by considering the \textit{time-augmented} version of the original path (see Definition \ref{def:sig_time}, Theorem \ref{thm:unique} and \ref{thm:uni_nonlinear} below).

\subsection{Signature of Discrete Data}

To define and compute signatures of discrete data streams, one can simply do linear interpolations and then apply signature transforms.
\begin{Definition}\label{def:data_stream}
The space of streams of data is defined as
\begin{equation}
    \mathcal{S}\left(\mathbb{R}^{d}\right)=\left\{\pmb{x}=\left(\pmb{x}_{1}, \ldots, \pmb{x}_{n}\right): \pmb{x}_{i} \in \mathbb{R}^{d}, n \in \mathbb{N}\right\}.\notag
\end{equation}
Given $\pmb{x}=\left(\pmb{x}_{1}, \ldots, \pmb{x}_{n}\right) \in \mathcal{S}\left(\mathbb{R}^{d}\right)$, the integer $n$ is called the length of $\pmb{x}.$ Furthermore for $a, b \in \mathbb{R}$ such that $a<b$, fix
\begin{equation}\label{eqn:linear_interpolate}
    a=u_{1}<u_{2}<\cdots<u_{n-1}<u_{n}=b.
\end{equation}
Let $X=\left(X^{1}, \ldots, X^{d}\right):[a, b] \rightarrow \mathbb{R}^{d}$ be continuous such that $X_{u_{i}}=\pmb{x}_{i}$ for all $i$, and linear on the intervals in between. Then $X$ is called a linear interpolation of $\pmb{x}$.
\end{Definition}

\begin{Definition}\label{def:sig_discrete}
Let $\pmb{x}=\left(\pmb{x}_{1}, \ldots, \pmb{x}_{n}\right) \in \mathcal{S}\left(\mathbb{R}^{d}\right)$ be a stream of data. Let $X$ be a linear interpolation of $\pmb{x}$. Then the signature of $\pmb{x}$ is defined as
$\operatorname{Sig}(\pmb{x})=\operatorname{Sig}(X),$ and the truncated signature of depth $N$ of $\pmb{x}$ is defined as $\operatorname{Sig}^{N}(\pmb{x})=\operatorname{Sig}^{N}(X)$.
\end{Definition}

\begin{Definition}\label{def:sig_time}
Given a path $X:[a, b] \rightarrow \mathbb{R}^{d}$, define the corresponding time-augmented path by $\widehat{X}_{t}=\left(t, X_{t}\right)$,  a path in $\mathbb{R}^{d+1} .$
\end{Definition}

\subsection{Key Properties of Signature}
\begin{Theorem}[Uniqueness \citep{hambly2010uniqueness}]\label{thm:unique} Let $X:[a, b] \rightarrow \mathbb{R}^{d}$ be a continuous piecewise smooth path. Then $\operatorname{Sig}(\widehat{X})$ uniquely determines $X$ up to translation.
\end{Theorem}

In fact, the signature not only determines a path uniquely up to translation, but also \textit{linearizes} any continuous functions of the path, as stated in the next theorem.

\begin{Theorem}[Universal nonlinearity \citep{arribas2018derivatives}]\label{thm:uni_nonlinear}
Let $F$ be a real-valued continuous function on continuous piecewise smooth paths in $\mathbb{R}^{d}$ and let $\mathcal{K}$ be a compact set of such paths. Furthermore assume that $X_{0}=0$ for all $X \in \mathcal{K}$. (To remove the translation invariance.) Let $\varepsilon>0$. Then there exists a linear functional $L$ such that for all $X \in \mathcal{K}$,
   $ |F(X)-L(\operatorname{Sig}(\widehat{X}))|<\varepsilon.$
\end{Theorem}

This \textit{universal nonlinearity} is the key property of the signature transform and is important for our model, and in general for applications in feature augmentations. See \cite{lyons2014feature}, \cite{li2017lpsnet}, \cite{morrill2020generalised} for examples.

Note that the signature by definition is an infinite-dimensional tensor. In practice, one can only compute the truncated signature $\operatorname{Sig}^N$  in  \eqref{eqn:sig_N} up to some depth $N$. The next result guarantees that reminder terms in the truncation decay factorially.

\begin{Theorem}[Factorial decay \citep{lyons2007differential}]\label{thm:fac_decay}
Let $X:[a, b] \rightarrow \mathbb{R}^{d}$ be a continuous piecewise smooth path and let  $\|\cdot\|$ be a tensor norm on $\left(\mathbb{R}^{d}\right)^{\otimes k}$.  Then
$$
\left\|\int_{a<t_{1}<\cdots<t_{k}<b} \mathrm{~d} X_{t_{1}} \otimes \cdots \otimes \mathrm{d} X_{t_{k}}\right\| \leq \frac{C(X)^{k}}{k !},
$$
where $C(X)$ is a constant depending on $X$.
\end{Theorem}

The next property about signatures, Theorem \ref{thm:inv_time_repara}, is the \textit{invariance to time reparameterizations}. It implies that the signature encodes the data by its arrival order and independently of its arrival time. This is a desired property in many applications such as hand-writing recognition \cite{yang2015chinese}, \cite{xie2017learning}. Meanwhile, there is an interesting interplay between Theorem \ref{thm:uni_nonlinear} and Theorem \ref{thm:inv_time_repara}: in a problem where time parameterizations are irrelevant, it suffices to compute the signature of $X$ by Theorem \ref{thm:inv_time_repara}; However, if time parameterization is important, then according to Theorem \ref{thm:uni_nonlinear}, applying the signature transform to the time-augmented path $\widehat{X}$ ensures that parameterization-dependent features are still learned.

\begin{Theorem}[Invariance to time reparameterizations \citep{lyons2007differential}]\label{thm:inv_time_repara}
Let $X:[0,1] \rightarrow \mathbb{R}^{d}$ be a continuous piecewise smooth path. Let $\psi:[0,1] \rightarrow[0,1]$ be continuously differentiable, increasing, and surjective. Then $\operatorname{Sig}(X)=\operatorname{Sig}(X \circ \psi)$.
\end{Theorem}

Note that by Theorem \ref{thm:inv_time_repara}, the signature of a stream of data is independent of the choice of $u_i$ in a linear interpolation in  \eqref{eqn:linear_interpolate}. Meanwhile, by Theorem \ref{thm:uni_nonlinear}, in order to learn parameterization-dependent features, one can apply the signature transform to the time-augmented data stream $\widehat{\pmb{x}}=\left(\widehat{\pmb{x}}_{1}, \ldots, \widehat{\pmb{x}}_{n}\right)$, where $\widehat{\pmb{x}}_{i}=(t_i, \pmb{x}_i)\in\mathbb{R}^{d+1}$, and $t_i$ is the time when the data point $\pmb{x}_i$ arrives.

Let $\pmb{x}=\left(\pmb{x}_{1}, \ldots, \pmb{x}_{n}\right) \in \mathcal{S}\left(\mathbb{R}^{d}\right)$ be a data stream of length $n$ in $\mathbb{R}^{d}.$ Then $\operatorname{Sig}^{N}(\pmb{x})$ has
\begin{equation}\label{eqn:sig_size}
    M(d,N):=\sum_{k=0}^{N} d^{k}=\frac{d^{N+1}-1}{d-1}
\end{equation}
components. In particular, the number of components does not depend on the length of the data stream $n$. The truncated signature maps the infinite-dimensional space of streams of data $\mathcal{S}\left(\mathbb{R}^{d}\right)$ into a finite-dimensional space of dimension $\left(d^{N+1}-1\right) /(d-1)$. Thus the signature is an efficient way to tackle long streams of data, or streams of variable length. 

\subsection{Computation of Signature Transform}\label{app:comp_sig}
The signature transform of a data stream can be computed in an efficient and tractable way, with the help of Chen's identity \citep{lyons2007differential,bonnier2019deep}. It starts by introducing the following $\boxtimes$ operation:
with $A_{0}=B_{0}=1$, define $\boxtimes$ by
\begin{align}\label{eqn:boxtimes}
&\boxtimes:\left(\prod_{k=1}^{N}\left(\mathbb{R}^{d}\right)^{\otimes k}\right) \times\left(\prod_{k=1}^{N}\left(\mathbb{R}^{d}\right)^{\otimes k}\right)  \rightarrow \prod_{k=1}^{N}\left(\mathbb{R}^{d}\right)^{\otimes k}, \notag\\
&\left(A_{1}, \ldots A_{N}\right) \boxtimes\left(B_{1}, \ldots, B_{N}\right)  \mapsto\left(\sum_{j=0}^{k} A_{j} \otimes B_{k-j}\right)_{1 \leq k \leq N} .
\end{align}
Chen's identity \citep{friz2010multidimensional} states that the image of the signature transform forms a group structure with respect to $\boxtimes$. That is, given a sequence of data $\left(x_{1}, \ldots, x_{L}\right) \in$ $\mathcal{S}\left(\mathbb{R}^{d}\right)$ and some $j \in\{2, \ldots, L-1\}$, 
\begin{equation*}
\operatorname{Sig}^{N}\left(\left(x_{1}, \ldots, x_{L}\right)\right)=\operatorname{Sig}^{N}\left(\left(x_{1}, \ldots, x_{j}\right)\right) \boxtimes \operatorname{Sig}^{N}\left(\left(x_{j}, \ldots, x_{L}\right)\right).    
\end{equation*}
Furthermore, from Example \ref{exp:sig_linear}, the signature of a sequence of length two can be computed explicitly from the definition. Letting
\begin{equation}\label{eqn:exp}
    \exp : \mathbb{R}^{d} \rightarrow \prod_{k=1}^{N}\left(\mathbb{R}^{d}\right)^{\otimes k}, \quad \exp : v \rightarrow\left(v, \frac{v^{\otimes 2}}{2 !}, \frac{v^{\otimes 3}}{3 !}, \ldots, \frac{v^{\otimes N}}{N !}\right),
\end{equation}
then
\begin{equation*}
    \operatorname{Sig}^{N}\left(\left(x_{1}, x_{2}\right)\right)=\exp \left(x_{2}-x_{1}\right).
\end{equation*}
Chen's identity further implies that the signature transform can be computed by
\begin{equation}\label{eqn:sig_compute}
    \begin{aligned}
        \operatorname{Sig}^{N}\left(\left(x_{1}, \ldots, x_{L}\right)\right)
         &=\exp \left(x_{2}-x_{1}\right) \boxtimes \exp \left(x_{3}-x_{2}\right)\\
        &\qquad \boxtimes \cdots \boxtimes \exp \left(x_{L}-x_{L-1}\right).
    \end{aligned}
\end{equation}
 \eqref{eqn:sig_compute} implies that computing the signature of an incoming stream of data is efficient and scalable. Indeed, suppose one has obtained a stream of data and computed its signature. Then after the arrival of some more data, in order to compute the signature of the entire signal, one only needs to compute the signature of the new piece of information, which is then computed via the tensor product with the previously-computed signature.

\paragraph{Improving computational efficiency.} Recall from  \eqref{eqn:sig_compute} that the signature may be computed by evaluating several $\boxtimes$ in  \eqref{eqn:boxtimes} and $\operatorname{exp}$ in  \eqref{eqn:exp}. We begin by noticing that the key component in the computation is to evaluate
\begin{equation*}\label{eqn:sig_compute_key}
    \left(\prod_{k=1}^{N}\left(\mathbb{R}^{d}\right)^{\otimes k}\right) \times \mathbb{R}^{d} \rightarrow \prod_{k=1}^{N}\left(\mathbb{R}^{d}\right)^{\otimes k}, \quad A, z \mapsto A \boxtimes \exp (z).
\end{equation*}
Instead of computing $A \boxtimes \exp (z)$ conventionally through the composition of $\exp$ and $\boxtimes$, \cite{kidger2020signatory} suggests to speed up the computation by Horner's method. More specifically, it is to expand
$$
A \otimes \exp (z)=\left(\sum_{i=0}^{k} A_{i} \otimes \frac{z^{\otimes(k-i)}}{(k-i) !}\right)_{1 \leq k \leq N},
$$
so that the $k$-th term can be computed by
\begin{equation*}\label{eqn:horner}
\begin{aligned}
     \sum_{i=0}^{k} A_{i} \otimes \frac{z^{\otimes(k-i)}}{(k-i) !} = \Bigg(&\Bigg(\cdots\left(\left(\frac{z}{k}+A_{1}\right) \otimes 
    \frac{z}{k-1}+A_{2}\right) \otimes \frac{z}{k-2}+\cdots\Bigg)\\
    &\otimes \frac{z}{2}+A_{k-1}\Bigg) \otimes z+A_{k}.
\end{aligned}
\end{equation*}

As proved in \cite{kidger2020signatory}, this method has uniformly (over $d, N$) fewer scalar multiplications than the conventional approach, and reduces the asymptotic complexity of this operation from $\mathcal{O}\left(N d^{N}\right)$ to $\mathcal{O}\left(d^{N}\right)$. Furthermore, this rate is asymptotically optimal, since the size of the result (an element of $\prod_{k=1}^{N}\left(\mathbb{R}^{d}\right)^{\otimes k}$), is itself of size $\mathcal{O}\left(d^{N}\right)$.

\section{A Generic Framework for Linear Statistical Models with Signature Transform}

The universal nonlinearity of the signature transform (cf. Theorem \ref{thm:uni_nonlinear}) suggests that linear models can effectively capture complex non-linear relationships between factors and targets. This section introduces a generic framework for time series forecasting that integrates linear statistical models with the signature transform, utilizing its capacity to encode intricate temporal patterns. Additionally, a weighting technique based on signature kernels \cite{sriperumbudur2010hilbert, berlinet2011reproducing, gretton2012kernel, kiraly2019kernels, chevyrev2022signature, salvi2021signature, issa2024non, pannier2024path, lemercier2024high} is incorporated into the linear models to enable adaptive weighting and regime switching.

\subsection{Forecasting Problem} 

Consider a time series forecasting problem involving two time series, \(\{\pmb{x}_\tau\}_{\tau \in \mathbb{N}^+}\) and \(\{y_\tau\}_{\tau \in \mathbb{N}^+}\). Here, \(\pmb{x}_\tau \in \mathcal{X} \subseteq \mathbb{R}^{d_0}\) is a \(d_0\)-dimensional vector representing the factor values available at time \(\tau\). The objective is to predict the value of \(y_{\tau + \Delta_t}\) for a given time interval \(\Delta_t \in \mathbb{N}^+\).
Thus, the goal is to find a (possibly nonlinear) model \(f^*_{\Delta_t} \in \mathcal{F} \subseteq \{f \mid f: \mathcal{X} \to \mathcal{Y}\}\) such that \(f^*_{\Delta_t}(\pmb{x}_\tau) \approx y_{\tau + \Delta_t}\), where \(\mathcal{F}\) is the class of all admissible models.

More specifically, given data up to time \(t\), \(\{(\pmb{x}_\tau, y_\tau)\}_{\tau = 1, 2, \dots, t}\), the goal is to predict \(y_{t + \Delta_t}\). A standard approach for finding \(f^*_{\Delta_t} \in \mathcal{F}\) is to solve the following optimization problem:
\[
f^*_{\Delta_t} \in \argmin_{f \in \mathcal{F}} \left\{ \frac{1}{t - \Delta_t} \sum_{\tau=1}^{t - \Delta_t} L\left(f(\pmb{x}_\tau), y_{\tau + \Delta_t}\right) \right\},
\]
where \(L: \mathcal{Y} \times \mathcal{Y} \to \mathbb{R}\) is a loss function that quantifies the discrepancy between the model prediction \(f(\pmb{x}_\tau)\) and the actual value \(y_{\tau + \Delta_t}\).
Once \(f^*_{\Delta_t}\) is obtained, the prediction for \(y_{t + \Delta_t}\) is given by \(\widehat{y}_{t + \Delta_t} := f^*_{\Delta_t}(\pmb{x}_t)\).


\subsection{Our Approach}
We present a generic framework that integrates linear statistical models, signature transform, and signature kernels.

\paragraph{Linear statistical models with signature transform.}   For any time step $\tau\in\N^+$ and time window size $l\in\N$, denote $\boldsymbol{x}_{\tau-l:\tau}:=(\boldsymbol{x}_{\tau-l}, \cdots, \boldsymbol{x}_{\tau})$ as the slice of the time series $\{\boldsymbol{x}_t\}_{t\in\N^+}$ from time $\tau-l$ to $\tau$.
Let $N$ be the depth of the truncated signature and $\Sig^N(\boldsymbol{x}_{\tau-l: \tau})$ be the depth-$N$ signature transform of the truncated path $\{x_\tau\}_{\tau\in \mathbb{N}^+}$ from time $\tau-l$ to $\tau$.

Let \(\Theta_{\text{linear}}\) denote a class of linear models that map \(\Sig^N(\boldsymbol{x}_{\tau-l: \tau})\) to \(y_{\tau + \Delta t}\). Given a training dataset \(\{(\pmb{x}_\tau, y_\tau)\}_{\tau=1}^t\), the optimal predictor \(\widehat{\pmb{\theta}}\) is determined by minimizing the empirical loss:
\begin{equation}\label{eq:linear_models}
    \widehat{\pmb{\theta}} \in \argmin_{\pmb{\theta} \in \mathbb{R}^d} \Bigg\{\frac{1}{t - \Delta t} \sum_{\tau=1}^{t - \Delta t} L_{\text{linear}} \Big(y_{\tau + \Delta t}, \Sig^N(\pmb{x}_{\tau-l: \tau}); \pmb{\theta} \Big) \Bigg\},
\end{equation}
where \(L_{\text{linear}}\) is the loss function, whose form depends on the specific choice of the linear statistical model. 
For example, in the case of ordinary least squares (OLS), the loss function \(L_{\text{linear}}\) is defined as the squared error between the actual and predicted values, i.e., 
\[L_{\text{OLS}} = \big(y_{\tau + \Delta t} - \Sig^N(\pmb{x}_{\tau-l: \tau}) \cdot \pmb{\theta} \big)^2.\] 
For Lasso regression, the loss function is augmented with an \(L_1\)-regularization term to promote sparsity in the parameters, resulting in 
\[L_{\text{LASSO}} = \big(y_{\tau + \Delta t} - \Sig^N(\pmb{x}_{\tau-l: \tau}) \cdot \pmb{\theta} \big)^2 + \lambda \|\pmb{\theta}\|_1,\] 
where \(\lambda > 0\) is the regularization parameter. Similarly, Ridge regression incorporates an \(L_2\)-regularization term to penalize large parameter values, leading to the loss 
\[L_{\text{Ridge}} = \big(y_{\tau + \Delta t} - \Sig^N(\pmb{x}_{\tau-l: \tau}) \cdot \pmb{\theta} \big)^2 + \lambda \|\pmb{\theta}\|_2^2.\]
The choice of the specific linear model and its associated loss function depends on the characteristics of the data and the desired trade-off between prediction accuracy and model parameter sparsity.

\paragraph{Adaptive weight via signature kernel.}

In the classical approach of linear statistical models in \eqref{eq:linear_models}, each historical sample is given equal weight in the optimization problem to obtain the model at time $t$. However, this equal-weight scheme may fail to account for changes of regime or seasonality. 
Instead, a more effective approach would be to dynamically assign weights based on their similarity to the current period. This is precisely what we propose, as elaborated below.

First, recall the signature feature map  (Theorem \ref{thm:uni_nonlinear}),
\begin{equation}\label{eqn:feature}
    \Phi: X \mapsto \text{Sig}(\widehat{X})
\end{equation}
is a universal feature map from the path space to the linear space of signatures \cite{chevyrev2022signature}. 
To avoid computation over a large space of functions, we  kernelize the signature feature map $\Phi$ in  \eqref{eqn:feature}, and define the signature kernel  
    $k(\pmb{a}, \pmb{b}):=\langle\Phi(\pmb{a}), \Phi(\pmb{b})\rangle$, 
for any discrete time series $\pmb{a}$ and $\pmb{b},$
as suggested in \cite{chevyrev2022signature}.
Here $\langle\cdot, \cdot\rangle$ is the inner product on the linear space of signatures. 

 Next, to measure the similarity between two discrete time series $\pmb{a}$ and $\pmb{b}$, consider the distance induced by the signature kernel \citep{sriperumbudur2010hilbert, berlinet2011reproducing, gretton2012kernel, kiraly2019kernels, chevyrev2022signature},
\begin{equation}\label{eqn:MMD_kernel_discrete}
    d_{\text{Sig}}(\pmb{a}, \pmb{b})=k\left(\pmb{a}, \pmb{a}\right)-2k(\pmb{a}, \pmb{b})+k\left(\pmb{b}, \pmb{b}\right).
\end{equation}
Small $d_{\text{Sig}}(\pmb{a}, \pmb{b})$ implies a higher similarity between patterns in $\pmb{a}$ and $\pmb{b}$, which suggests that $\pmb{a}$ and $\pmb{b}$ come from the same regime and share the similar seasonality. In practice, one may truncate the signature to depth $N$ when computing  \eqref{eqn:MMD_kernel_discrete}, and we denote the distance computed from the depth-$N$ truncated signature by $d_{\text{Sig}, N}$.

 Finally, we adapt the weights in the linear models according to the signature kernel, this is called $\mathtt{AdaWeight.Sig}$. It takes five hyper-parameters: a forecast horizon $\Delta t$,  a window size $l\in\N$, a signature depth $N\in\N^+$, a temperature parameter $\gamma\geq 0$, and the distance metric $d_{\text{Sig}, N}$. 
More precisely,
define $\pmb{z}_\tau:=(\pmb{x}_\tau, y_{\tau +  \Delta{t}})$ for any $\tau\in\N^+$;
for any $\tau_1,\tau_2\in\N^+$ and $\tau_1<\tau_2$, denote $\pmb{z}_{\tau_1:\tau_2}:=(\pmb{z}_{\tau_1}, \cdots, \pmb{z}_{\tau_2})$ as a slice of the time series $\{\pmb{z}_t\}_{t\in\N^+}$ from time $\tau_1$ to $\tau_2$. 
At each time $t$, $\mathtt{AdaWeight.Sig}$ takes the historical samples  $\{(\pmb{x}_\tau, y_{\tau+{ \Delta t}})\}_{  1 \leq \tau \leq t-\Delta t}$ as input, and outputs an adaptive weight vector 
\begin{equation}\label{eqn:adaptive_weight}
    \pmb{W}^{ (\Delta t)}:=(w_{1}^{({  \Delta t})}, w_{2}^{({  \Delta t})}, \cdots, w_{t-{  \Delta t}}^{({  \Delta t})})\in\R^{t-{  \Delta t}}_{\geq 0},
\end{equation}
with $\sum_{\tau=1}^{t-{  \Delta t}}w^{({  \Delta t})}_{\tau}=1$. That is, $\mathtt{AdaWeight.Sig}$ will assign a higher weight $w_{\tau}^{(\Delta t)}$ to the sample $(\pmb{x}_\tau, y_{\tau+{  \Delta t}})$ if the seasonality pattern near time $\tau$ is more similar to the current seasonality pattern embedded in the sample $(\pmb{x}_{t - \Delta t}, y_{t})$. 
Thus, when a new data sample arrives,
the weight vector $\pmb{W}^{({  \Delta t})}$ will be recomputed by the $\mathtt{AdaWeight.Sig}$ module, to adapt to changes in the recent data samples.

\begin{algorithm}
  \caption{\textbf{Adaptive Weight via Signature Kernel ($\mathtt{AdaWeight.Sig}$)}}
  \label{algo:sig_id}
\begin{algorithmic}[1]
    \STATE \textbf{Input}: forecast horizon ${  \Delta t}$, window size $l$, signature depth $N$, temperature parameter $\gamma$, kernel-based distance metric $d_{\text{Sig}, N}$ \eqref{eqn:MMD_kernel_discrete}, data set $D= \{\pmb{z}_\tau =(\pmb{x}_\tau, y_{\tau+{  \Delta t}})\}_{1 \leq \tau\leq t-{  \Delta t}}$.
    \FOR {$\tau=1,2,\cdots,t-{  \Delta t}$}
        \STATE Compute the distance $\delta_\tau$ between the truncated depth-$N$ signatures of $\pmb{z}_{\tau-l:\tau}$ and $\pmb{z}_{t -{  \Delta t}-l:t-{  \Delta t}}$:
        \begin{equation}\label{eqn:sig_dist}
            \delta_\tau:=d_{\text{Sig}, N}\left(\pmb{z}_{\tau-l:\tau}, \pmb{z}_{t-{  \Delta t}-l:t-{  \Delta t}}\right).
        \end{equation}
    \ENDFOR
    \FOR {$\tau=1,2,\cdots,t-{  \Delta t}$}
        \STATE Compute the weight $w_\tau^{(\Delta t)}$ according to
        \begin{equation}
            w_\tau^{(\Delta t)}:=\frac{\exp(-\gamma\cdot\delta_\tau)}{\sum_{\tau=1}^{t-{  \Delta t}} \exp(-\gamma\cdot\delta_\tau)}.
        \end{equation}
    \ENDFOR
    \STATE \textbf{Output}: the weight vector $\pmb{W}^{({  \Delta t})} =(w_1^{({  \Delta t})},\cdots, w_{t-{  \Delta t}}^{({  \Delta t})})$.
\end{algorithmic}
\end{algorithm}

 \paragraph{Adaptive regression via signature kernel.}
 Measuring similarity via signature kernel leads to a novel linear modeling approach, in which we propose to adapt weights according to the similarities between time series data to capture seasonality and regime switching embedded in the data. 
In the case of forecasting models with signature transforms, comparing similarities of data translates to identifying  ``signature feature maps''. This is the statistical equivalence of identifying different distributions for a given set of data, albeit in the linearized features space from the signature transform. 
Algorithm \ref{algo:lasso_sig} summarizes this 
 approach of adaptive linear models via signature kernel, by integrating $\mathtt{AdaWeight.Sig}$ outlined in Algorithm \ref{algo:sig_id} with the linear modeling in \eqref{eq:linear_models}.

\begin{algorithm}
  \caption{\textbf{Adaptive linear models with Signature Kernel}}
  \label{algo:lasso_sig}
\begin{algorithmic}[1]
    \STATE \textbf{Input}: regularization parameter $\lambda$, window size $l$, signature depth $N$, temperature parameter $\gamma$,  kernel-based distance metric $d_{\text{Sig}, N}$, data set $D= \{\pmb{z}_\tau =(\pmb{x}_\tau, y_{\tau+{  \Delta t}})\}_{1 \leq \tau\leq t-{  \Delta t}}$.
    \FOR {${\Delta t}=1, 2, \cdots, \Delta T$}

        \STATE \textbf{Regime identification}: Run $\mathtt{AdaWeight.Sig}$ (Algorithm \ref{algo:sig_id})  with $\Delta t$, $l$, $N$,  $\gamma$,  $d_{\text{Sig}, N}$, and $D$. Output $\pmb{W}^{(\Delta t)}$ in \eqref{eqn:adaptive_weight}. 

        \STATE \textbf{Adaptive regression}: Apply the linear model on the data set $D$, with adaptive weight $\pmb{W}^{({  \Delta  t})}$ and regularization parameter $\lambda$: 
        \begin{align}\label{eqn:lasso_weight} 
            \widehat{\boldsymbol{\theta}}_{{  \Delta  t}}^{\lambda} \in \argmin_{\boldsymbol{\theta}\in\R^d}\Bigg\{\sum_{\tau=1}^{t-{  \Delta  t}} w^{({  \Delta  t})}_{\tau}\cdot L_{\text{linear}} \left(y_{\tau + \Delta t}, \Sig^N(\boldsymbol{x}_{\tau-l:\tau});\boldsymbol{\theta}\right)\Bigg\}. 
           \end{align}
        
        \STATE Given $\pmb{x}_t, y_t$, make prediction on $y_{t + {  \Delta  t}}$:
        $$\widehat{y}_{t+{  \Delta  t}}= \text{Sig}^N(\boldsymbol{x}_{t-l:{ t}})\cdot{\widehat{\boldsymbol{\theta}}}_{{  \Delta  t}}^{\lambda}.$$
    \ENDFOR
    \STATE \textbf{Output:} the forecast $ (\widehat{y}_{t+1}, \cdots, \widehat{y}_{t + \Delta T}).$
\end{algorithmic}
\end{algorithm}


\section{Forecasting Problem of Transportation Marketplace rate}\label{sec: transporation_solution} 
The freight marketplace rate forecast problem involves two time series $\{\pmb{x}_\tau\}_{\tau\in\N^+}$  and $\{y_\tau\}_{\tau\in\N^+}$. Here,  $\pmb{x}_\tau\in\Xc\subseteq\R^{d_0}$ is a $d_0$-dimensional vector representing values of the key economic factors that drive the supply and demand in the freight marketplace at time $\tau$.
Factors from the market supply side include information regarding the supply of drivers and trucks and fuel/oil prices. Market demand factors include imports, agriculture information, manufacturing activities, housing indexes, and railway transport. Additionally, $y_\tau \in \Yc \subseteq \mathbb{R}$ is the freight marketplace rate at time $\tau$.
Previously, Amazon relied on a commercial service to obtain forecasts for future marketplace rates. However, those forecasts lacked accuracy and transparency. To address these, we consider the following forecast problem.

\subsection{Our Approach}
We will present a signature-based adaptive two-step LASSO approach that we have developed and implemented in Amazon, which has demonstrated excellent performance in solving this problem. 
 
\paragraph{Data.} Our approach and experiment start by collecting data involving over a hundred of national and regional market supply and demand factors, downloaded from the governmental public websites, including Federal Reserve Bank and Bureau of Labor Statistics, as well as industrial databases such as Logistic Manager. The time range for the data is from 2018 to 2022.  

\paragraph{External factor preprocessing.} We first run a correlation test between the marketplace rate and the factors to remove non-significant factors, with further correlation analysis to identify factors that may be colinear.  After this round of elimination, over forty factors remain, including consumer price index,  housing index, oil and gas drilling, logistic managers' index, employment information, weather, and other market benchmarks.

\paragraph{Internal features via signature transform.} Besides the ``external''  factors $\pmb{x}_\tau$, most time series forecasting approaches, such as ARIMA, also construct ``internal'' features from the history of $y_\tau$. Those ``internal'' features may help to characterize the trend, momentum, and stationarity of $y_\tau$. We, instead, exploit the universal nonlinearity property of signature transform (Theorem \ref{thm:uni_nonlinear}), and construct signature features as suitable candidates for the ``internal'' features. {In the application to transportation rates, given the extensive dataset of external factors, preliminary experiments indicate that it is sufficient to consider the signature transform of \(y\) alone, rather than incorporating both \(\boldsymbol{x}\) and \(y\), as in \eqref{eq:linear_models}.}
\begin{figure}
    \centering
    \includegraphics[width=0.6\textwidth]{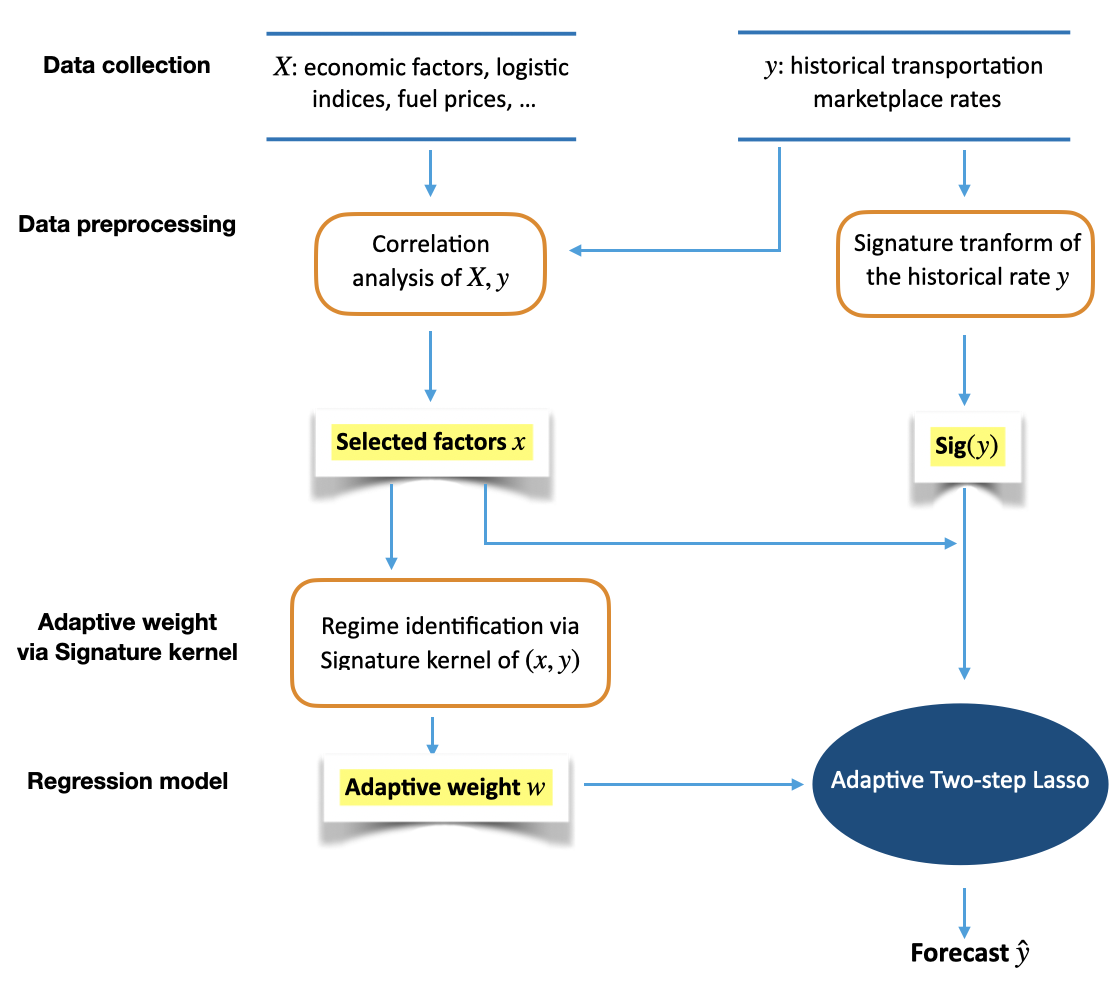}
 \caption{Flowchart illustrating the application of Algorithm \ref{algo:lasso_sig} to transportation marketplace rate forecast}
    \label{fig:workflow}
\end{figure}
More specifically, for any time step $\tau\in\N^+$ and time window size $l\in\N$, denote $y_{\tau-l:\tau}:=(y_{\tau-l}, \cdots, y_{\tau})$ as the slice of the time series $\{y_t\}_{t\in\N^+}$ from time $\tau-l$ to $\tau$. The feature vector for predicting $y_{\tau + {  \Delta t}}$ consists of both the economic factors $\pmb{x}_\tau$ and the depth-$N$ signature features $\text{Sig}^N(y_{\tau-l:\tau})$. We denote the concatenation of those two sets of features as $\left[\boldsymbol{x}_\tau, \text{Sig}^N(y_{\tau-l:\tau})\right]$, whose dimension is denoted by $d$.

\paragraph{Two-step LASSO.}

The universal nonlinearity property of the signature transform linearizes the feature space, hence translating the forecasting problem into a linear regression analysis. 
Since the dimension $d$ of the feature vector may be relatively large compared to the number of historical samples, especially when the time step $t$ is small, we adopt the approach of two-step LASSO to avoid overfitting and the issue of co-linearity especially between the signature feature and the external factors.   
The first step is to select the factors by solving the standard LASSO regression \cite{tibshirani1996regression}, \cite{zhao2006model}, \cite{zou2006adaptive}. This is to add an $L_1$-regularization to model coefficients in the ordinary least square objective. This $L_1$-regularization will encourage the sparsity of model coefficients, and prevent the over-fitting problem. In the second step, an OLS with only the selected factors is applied. 
This two-step LASSO estimation procedure has been shown to produce a smaller bias than standard LASSO for a range of models \cite{belloni2013least}, \cite{chetelat2017optimal}.

More precisely, let $W^{(\Delta t)} = (w_1^{\Delta t},\cdots w_{t-\Delta t}^{(\Delta t)})$ denote the adaptive weight vector in \eqref{eqn:adaptive_weight}, which is the output of Algorithm \ref{algo:sig_id}. The adaptive LASSO regression is to solve the following optimization problem:
    \begin{align}
\widehat{\boldsymbol{\theta}}_{\text{LASSO},{  \Delta  t}}^{\lambda}\in&\argmin_{\boldsymbol{\theta}\in\R^d} \Bigg\{ \sum_{\tau=1}^{t-{  \Delta t}} w_\tau^{(\Delta t)} \cdot \Big(y_{\tau+{  \Delta t}}
  -\left[\boldsymbol{x}_\tau, \text{Sig}^N(y_{\tau-l:\tau})\right]\cdot\boldsymbol{\theta}\Big)^2+\lambda \|\boldsymbol{\theta}\|_1\Bigg\}.\label{eqn:lasso}
    \end{align}
Here the constant $\lambda$, called the regularization parameter, controls the sparsity of coefficients:
a higher value of $\lambda$ leads to a smaller number of nonzero coefficients in $\widehat{\boldsymbol{\theta}}^{\lambda}_{\text{LASSO},   \Delta t}$. 
In the two-step LASSO,  the first step is to select the factors by solving the LASSO regression in  \eqref{eqn:lasso}, and get $\widehat{\boldsymbol{\theta}}^{\lambda}_{\text{LASSO},   \Delta t}$. In the second step, 
the subsequent OLS refitting is to find $\widehat{\boldsymbol{\theta}}^{\lambda}_{\text{two-step},   \Delta t}$ such that
\begin{align}\label{eqn:2step_lasso}
{\widehat{\boldsymbol{\theta}}}_{\text{two-step},{  \Delta  t}}^{\lambda} \in 
\argmin _{\text{supp}\left[\boldsymbol{\theta}\right]=\text{supp}\left[{\widehat{\boldsymbol{\theta}}}_{\text{LASSO},{  \Delta  t}}^{\lambda}\right]} 
\Bigg\{\sum_{\tau=1}^{t-{  \Delta t}} w_\tau^{(\Delta t)} \cdot \Big(y_{\tau+{  \Delta t}}-\left[\boldsymbol{x}_\tau, \text{Sig}^N(y_{\tau-l:\tau})\right]\cdot\boldsymbol{\theta}\Big)^2\Bigg\}. 
\end{align}
Figure \ref{fig:workflow} illustrates the overall workflow for predicting transportation marketplace rates using Algorithm \ref{algo:lasso_sig} with the two-step LASSO model.

\begin{table*}
\caption{Comparing national-level 3-month-ahead industry predictions with our model predictions}
    \begin{tabularx}{1.0\textwidth}{@{}>{\bfseries} p{0.1\textwidth}*{8}{X}@{}}
    \toprule
    \textbf{Month \newline{} in 2021} & \textbf{Actual}  & \textbf{Industrial\newline{} prediction} & \textbf{\% Error \newline{} Ind. prediction} & \textbf{Our prediction}  & \textbf{\% Error \newline{} Our prediction} & \textbf{\% Accuracy improvement}\\
    \midrule
    \midrule
    Apr  & \$2.44 & \$1.89 & 23\%  & \$2.39 & 2\%  &1050 \%\\
    May  & \$2.51 & \$1.82 & 27\%  & \$2.45 & 2\%   & 1250\%\\
    Jun  & \$2.53 & N/A   & N/A   & \$2.48 & 2\%    & {N/A}\\
    Jul  & \$2.57 & \$2.18 & 15\%  & \$2.53 & 2\%  & 650\% \\
    Aug  & \$2.61 & \$2.21 & 15\%  & \$2.58 & 1\%    & 1400\% \\
    Sep  & \$2.71 & \$2.23 & 18\%  & \$2.70 & 1\%  & 1700\% \\
    Oct  & \$2.72 & \$2.36 & 13\%  & \$2.69 & 1\%   & 1200\%\\
    Nov  & \$2.72 & \$2.38 & 13\%  & \$2.71 & 1\%  & 1200\%\\
    \bottomrule
    \end{tabularx}
       \label{tab:dat_vs_rfi}%
\end{table*}

\begin{table*}[htbp]
  \centering
   \caption{12-week ahead model predictions across different regions.}
    \begin{tabularx}{1.0\textwidth}{@{}>{\bfseries} p{0.1\textwidth}*{7}{X}@{}}
    \toprule
    \textbf{\# of week ahead  prediction} & \textbf{N.A.} & \textbf{A} & \textbf{B} & \textbf{C} & \textbf{D} & \textbf{E} \\
    \midrule
    \midrule
    \textbf{1} & 1.06\% & 1.88\% & 1.53\% & 1.38\% & 1.26\% & 1.53\% \\
    \textbf{2} & 1.53\% & 1.93\% & 2.76\% & 1.92\% & 2.08\% & 2.06\% \\
    \textbf{3} & 1.55\% & 2.95\% & 2.18\% & 1.19\% & 2.21\% & 2.07\% \\
    \textbf{4} & 1.61\% & 3.83\% & 1.87\% & 1.50\% & 2.67\% & 2.34\% \\
    \textbf{5} & 1.33\% & 2.57\% & 2.73\% & 1.05\% & 2.99\% & 1.32\% \\
    \textbf{6} & 1.44\% & 2.86\% & 2.96\% & 1.85\% & 2.78\% & 1.20\% \\
    \textbf{7} & 1.64\% & 2.33\% & 5.69\% & 1.73\% & 2.77\% & 2.51\% \\
    \textbf{8} & 1.86\% & 1.71\% & 5.25\% & 3.64\% & 2.55\% & 2.64\% \\
    \textbf{9} & 1.88\% & 2.99\% & 4.86\% & 2.55\% & 3.69\% & 2.15\% \\
    \textbf{10} & 2.30\% & 3.80\% & 3.31\% & 4.10\% & 2.13\% & 2.53\% \\
    \textbf{11} & 2.60\% & 4.71\% & 5.54\% & 4.33\% & 2.06\% & 2.85\% \\
    \textbf{12} & 2.38\% & 5.65\% & 5.25\% & 4.76\% & 2.74\% & 5.33\% \\
    \bottomrule
    \end{tabularx}%
  \label{tab:test_error}
\end{table*}%

\begin{table*}[htbp]
  \centering
  \caption{Comparison between predictions with and without
adaptive signature kernel}
    \begin{tabularx}{1.0\textwidth}{@{}>{\bfseries} p{0.1\textwidth}*{6}{X}@{}}
    \toprule
    \textbf{Week} & \textbf{Actual} & \multicolumn{1}{p{5em}}{\textbf{Prediction without signature kernel}} & \multicolumn{1}{p{6em}}{\textbf{\% Error without signature kernel}} & \multicolumn{1}{p{5em}}{\textbf{Prediction with signature kernel}} & \multicolumn{1}{p{6em}}{\textbf{\% Error with signature kernel}} \\
    \midrule
    \midrule
    \textbf{10/31/21} & 3.37  & 3.30  & 2.31\% & 3.32  & 1.53\% \\
    \textbf{11/7/21} & 3.45  & 3.27  & 5.19\% & 3.34  & 3.22\% \\
    \textbf{11/14/21} & 3.46  & 3.25  & 5.88\% & 3.38  & 2.15\% \\
    \textbf{11/21/21} & 3.52  & 3.22  & 8.65\% & 3.37  & 4.25\% \\
    \textbf{11/28/21} & 3.48  & 3.16  & 9.28\% & 3.35  & 3.66\% \\
    \textbf{12/5/21} & 3.49  & 3.15  & 9.85\% & 3.31  & 5.15\% \\
    \bottomrule
    \end{tabularx}%
      \label{tab:compare_seasonality}
\end{table*}%

\begin{table*}[htbp]
  \centering
   \caption{Our model  predictions posted on June 25 2022}
    \begin{tabularx}{1.0\textwidth}{@{}>{\bfseries} p{0.1\textwidth}*{7}{X}@{}}
    \toprule
    \multicolumn{1}{p{5em}}{\textbf{Region}} & \textbf{Week} & \multicolumn{1}{p{5em}}{\textbf{Actual}} & \multicolumn{1}{p{5em}}{\textbf{Prediction}} & \multicolumn{1}{p{5em}}{\textbf{\% Error}} \\
    \midrule
    \midrule
    \multicolumn{1}{c}{\multirow{5}[2]{*}{\textbf{N.A.}}} & \textbf{7/3/2022} & 1.95  & 1.93  & 1.34\% \\
          & \textbf{7/10/2022} & 1.95  & 1.93  & 1.12\% \\
          & \textbf{7/17/2022} & 1.94  & 1.95  & 0.37\% \\
          & \textbf{7/24/2022} & 1.93  & 1.96  & 1.25\% \\
          & \textbf{7/31/2022} & 1.94  & 1.95  & 0.90\% \\
    \midrule
    \multicolumn{1}{c}{\multirow{5}[2]{*}{\textbf{A}}} & \textbf{7/3/2022} & 1.8   & 1.78  & 0.68\% \\
          & \textbf{7/10/2022} & 1.8   & 1.76  & 2.06\% \\
          & \textbf{7/17/2022} & 1.76  & 1.76  & 0.04\% \\
          & \textbf{7/24/2022} & 1.72  & 1.74  & 1.46\% \\
          & \textbf{7/31/2022} & 1.66  & 1.73  & 4.03\% \\
    \midrule
    \multicolumn{1}{c}{\multirow{5}[2]{*}{\textbf{B}}} & \textbf{7/3/2022} & 1.26  & 1.26  & 0.39\% \\
          & \textbf{7/10/2022} & 1.21  & 1.27  & 4.68\% \\
          & \textbf{7/17/2022} & 1.21  & 1.25  & 3.44\% \\
          & \textbf{7/24/2022} & 1.21  & 1.25  & 3.28\% \\
          & \textbf{7/31/2022} & 1.24  & 1.25  & 1.25\% \\
    \midrule
    \multicolumn{1}{c}{\multirow{5}[2]{*}{\textbf{C}}} & \textbf{7/3/2022} & 1.1   & 1.1   & 0.40\% \\
          & \textbf{7/10/2022} & 1.08  & 1.09  & 0.70\% \\
          & \textbf{7/17/2022} & 1.02  & 1.08  & 5.32\% \\
          & \textbf{7/24/2022} & 1.02  & 1.04  & 2.01\% \\
          & \textbf{7/31/2022} & 1.02  & 1.01  & 0.60\% \\
    \midrule
    \multicolumn{1}{c}{\multirow{5}[2]{*}{\textbf{D}}} & \textbf{7/3/2022} & 1.92  & 1.95  & 1.86\% \\
          & \textbf{7/10/2022} & 1.94  & 1.96  & 0.99\% \\
          & \textbf{7/17/2022} & 1.98  & 1.98  & 0.28\% \\
          & \textbf{7/24/2022} & 2.01  & 2.03  & 1.06\% \\
          & \textbf{7/31/2022} & 2.01  & 2.07  & 2.86\% \\
    \midrule
    \multicolumn{1}{c}{\multirow{5}[2]{*}{\textbf{E}}} & \textbf{7/3/2022} & 1.79  & 1.76  & 1.70\% \\
          & \textbf{7/10/2022} & 1.76  & 1.75  & 0.75\% \\
          & \textbf{7/17/2022} & 1.76  & 1.74  & 0.94\% \\
          & \textbf{7/24/2022} & 1.76  & 1.73  & 1.89\% \\
          & \textbf{7/31/2022} & 1.75  & 1.73  & 1.01\% \\
    \bottomrule
    \end{tabularx}%
  \label{tab:prediction_2022}%
\end{table*}%

 \section{Real-time performance}
Our signature-based adaptive two-step LASSO algorithm has been implemented for the trucking operations in Amazon since November 2022.  Performance analysis shows that our forecast model presents superior performance than commercially available forecast models, with prediction accuracy improvement by more than fivefold, and has an estimated annualized savings of \$50 million. 

Below we will present the real-time performance using data from April 2021 to July 2022. While this timeframe precedes the model's actual implementation due to confidentiality restrictions, it represents a particularly challenging period marked by both the COVID-19 pandemic and the Ukraine conflict, and allows us to demonstrate the model's effectiveness in volatile market conditions.

We will showcase the national-level prediction in North America (N.A.) along with five representative regions within North America, designated A, B, C, D, and E to protect proprietary information. We apply the relative error to measure model performances where 
$\text{relative error} \coloneqq \frac{|\text{actual rate} - \text{forecast rate}|}{|\text{actual rate}|}.$

 \paragraph{Predictions comparison between industry models and our model.}
We compare the performance of our model at the national-level predictions with the standard industry predictions for the April 2021 - November 2021 time period. Both our model predictions and industry predictions are made three months (twelve weeks) ahead of time, with monthly predictions obtained by aggregating weekly predictions. The detailed results are listed in Table \ref{tab:dat_vs_rfi}. 
In particular, our predictions (with a relative error of around 2\%) are far superior to standard industry predictions (with a relative error of around 20\%). The prediction accuracy is improved by more than fivefold. 

\paragraph{Relative prediction errors up to twelve weeks.}
Table \ref{tab:test_error} reports the relative prediction error of our model for the national level and five regional predictions (A, B, C, D, and E), up to a twelve-week horizon.   The prediction error moderately increases from around $1\%$ for the one-week prediction to approximately $5\%$ for the twelve-week prediction, remaining significantly lower than the industry standard of $15\%$.

\paragraph{Necessity of adaptive signature kernel.}
To demonstrate the necessity of the adaptive signature kernel,  the key and novel component of our model,  we compare the predictions from Algorithm \ref{algo:lasso_sig} with and without the signature kernel in  \eqref{eqn:2step_lasso}. The results are reported in Table \ref{tab:compare_seasonality}. The predictions presented here are for one representative region in North America on October 24, 2021.   Evidently from Table \ref{tab:compare_seasonality},   the errors without the signature kernel are larger ($>8\%$) even for short-term predictions. In contrast, the signature kernel method captures better the seasonality, and obtains a smaller relative error ($<5\%$) for short-term predictions. 
This table shows the effectiveness of incorporating the signature kernel in the forecast model.

 \paragraph{Short-term prediction error.}
Table \ref{tab:prediction_2022} reports the relative prediction error of our model for the national level and five regional predictions (A, B, C, D, and E), for a five-week horizon, with model predictions made on June 25, 2022.   Most of the prediction errors are shown to be less than  $2\%$.

\section{Conclusion}
This work presents a novel, highly accurate signature-based adaptive two-step LASSO forecasting model for transportation marketplace rates. Deployed at Amazon since November 2022, it delivers more than fivefold forecast accuracy improvements compared to industry models even amidst major market disruptions, with significant cost savings. 


\bibliographystyle{alpha}
\balance
\bibliography{refs}

\end{document}